\documentclass[letterpaper, 10 pt, conference]{ieeeconf}  

\IEEEoverridecommandlockouts                              

\usepackage{graphics} 

\usepackage{epsfig} 

\usepackage{mathptmx} 

\usepackage{times} 

\usepackage{amsmath} 

\usepackage{amssymb}  
\usepackage{comment}
\usepackage{subfigure}
\usepackage{multirow}
\usepackage{diagbox}
\usepackage{booktabs}
\usepackage{cite}
\usepackage{color}
\makeatletter
\let\NAT@parse\undefined
\makeatother
\usepackage[pagebackref,colorlinks=true]{hyperref}

\title{\LARGE \bf
CoLRIO: LiDAR-Ranging-Inertial Centralized State Estimation for Robotic Swarms
}

\author{Shipeng Zhong, Hongbo Chen, Yuhua Qi$^{\star}$, Dapeng Feng, \\ Zhiqiang Chen, Jin Wu, Weisong Wen and Ming Liu 
	\thanks{This research was funded by Research on Path Planning Algorithm of Swarm Unmanned System Based on Deep Reinforcement Learning of China University Industry, Education and Research Innovation Fund No. 2021ZYA11010. Corresponding author is Yuhua Qi.}
	\thanks{Shipeng Zhong, Hongbo Chen, Yuhua Qi, Dapeng Feng and Zhiqiang Chen are with the School of Systems Science and Engineering, Sun Yat-sen University, Guangzhou 510006, China (e-mail: {zhongshp5, fengdp5, chenzq56}@mail2.sysu.edu.cn, {chenhongbo, qiyh8}@mail.sysu.edu.cn).
		{\tt\small }}%
	\thanks{J. Wu and M. Liu are with Department of Electronic and Computer Engineering, Hong Kong University of Science and Technology, Hong Kong (e-mail: jin\_wu\_uestc@hotmail.com, eelium@ust.hk).}
	\thanks{W. Wen is with Department of Aeronautical and Aviation Engineering, Hong Kong Polytechnic University, Hong Kong (e-mail: Welson.wen@polyu.edu.hk).}
}

\begin{document}
\maketitle
\thispagestyle{empty}
\pagestyle{empty}

\begin{abstract}
Collaborative state estimation using different heterogeneous sensors is a fundamental prerequisite for robotic swarms operating in GPS-denied environments, posing a significant research challenge.
In this paper, we introduce a centralized system to facilitate collaborative LiDAR-ranging-inertial state estimation, enabling robotic swarms to operate without the need for anchor deployment.
The system efficiently distributes computationally intensive tasks to a central server, thereby reducing the computational burden on individual robots for local odometry calculations.
The server back-end establishes a global reference by leveraging shared data and refining joint pose graph optimization through place recognition, global optimization techniques, and removal of outlier data to ensure precise and robust collaborative state estimation.
Extensive evaluations of our system, utilizing both publicly available datasets and our custom datasets, demonstrate significant enhancements in the accuracy of collaborative SLAM estimates. Moreover, our system exhibits remarkable proficiency in large-scale missions, seamlessly enabling ten robots to collaborate effectively in performing SLAM tasks. In order to contribute to the research community, we will make our code open-source and accessible at \url{https://github.com/PengYu-team/Co-LRIO}.

\end{abstract}
\section{Introduction}\label{sec:introduction}

Autonomous robotic swarms have gained significant traction in various public applications, such as exploration, inspection, search-and-rescue, healthcare, housework, and logistics \cite{multi_robot_system_survey:2019:rizk, cslamreview:2021:lajoie}.
In these applications, accurate estimation the six-degree-of-freedom states of individual robots is pivotal, as it underpins feedback control, obstacle avoidance, and path planning.
While state estimation and mapping for single-robot systems can leverage state-of-the-art SLAM techniques, such as Visual Inertial Odometry (VIO) \cite{vins:2018:qin, orb_slam3:2021:campos} and LiDAR Inertial Odometry (LIO) \cite{lio_sam:2020:shan, fast_lio2:2022:xu, Nguyen_lrio}, the transition to cooperative localization for robotics swarms introduces unique complexities, relying on onboard sensors and intra-swarm communication.

With the growing need for cooperative localization, various algorithms have been developed, some relying on external devices like motion capture \cite{approach:2008:jaimes}, fixed Ultra-WideBand (UWB) anchors \cite{robot:2015:ledergerber}, or Global Position System (GPS) \cite{outdoor_swarm:2016:moon} to estimate robot states in a common reference frame.
However, these solutions are not always available, often constrained by factors such as urban environments limiting GPS reliability or the need for infrastructure deployment (e.g., UWB anchors), restricting their versatility.

An alternative approach to circumvent these limitations involves establishing constraints between robots using vision detection or place recognition \cite{vio_uwb_based:2022:queralta, omni_swarm:2022:xu}.
These constraints can arise when sensors on multiple robots simultaneously observe shared landmarks or when one robot observes another within its sensor's field of view. However, shared observations necessitate significant view overlap between neighboring robots, constraining them to remain close and limiting the swarm's flexibility.

\begin{figure}[t]
	\centering
	\includegraphics[width=0.99\linewidth]{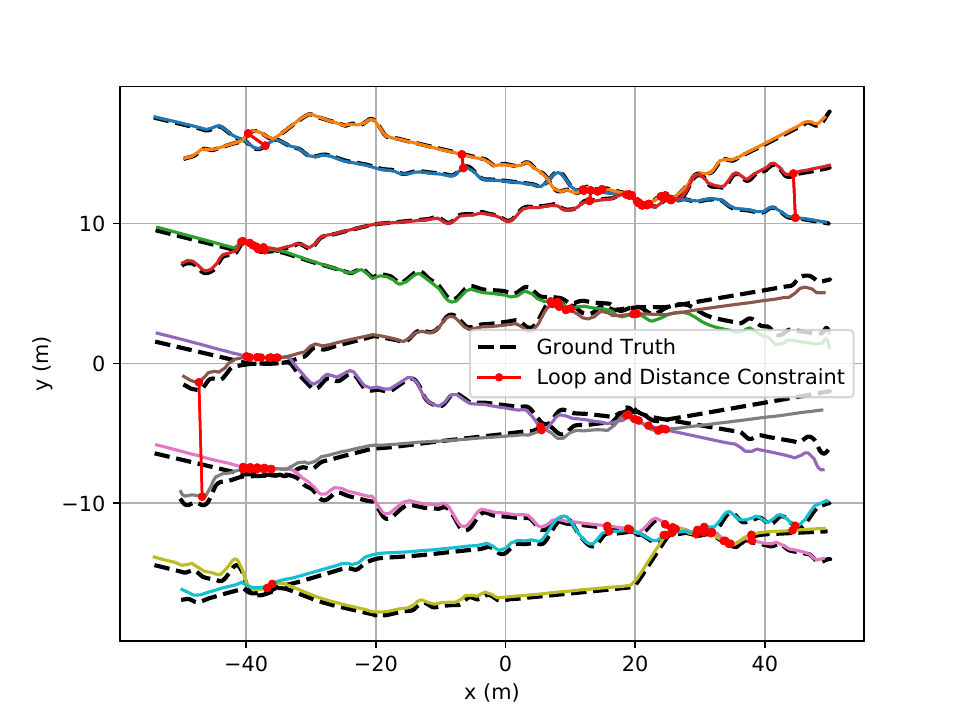}
	\caption{Simulation results of CoLRIO estimate from 10 drones. The total trajectory length exceeds 1000 m; the black dotted line indicates the ground truth, and the loop and distance constraints are denoted by red lines.}
	\label{fig:sim10traj}
	\vspace{-0.3cm} 
\end{figure}

This paper introduces a real-time and robust centralized state estimation system tailored for a team of robots equipped with LiDAR, IMU, and UWB sensor suites, enabling collaborative localization without deploying any anchors.
Each robot employs direct LIO using the intensity-aided Fast-GICP method for efficient scan-to-map registration.
Concurrently, a central server gathers pairwise distance measurements, conducts inter-robot loop closure detection, carries out robust joint optimization, and subsequently furnishes the robots with optimized poses as feedback.
Extensive performance evaluations of our system are conducted using publicly available datasets and our custom datasets, which reveal improvements in the accuracy of collaborative SLAM estimates.
Notably, the proposed system exhibits support for collaborative SLAM involving up to 10 robots (Fig. \ref{fig:sim10traj}), underlining its efficacy in swarm missions.

To summarize, the contributions of this work include:
({\romannumeral 1}): A tightly coupled LiDAR inertial odometry built atop fixed-lag smoothing and intensity-aided direct raw points registration.
({\romannumeral 2}): An online, robust, and centralized LiDAR-ranging-inertial state estimation system for the robotic swarm enables co-localization without deploying anchors.
({\romannumeral 3}): Extensive evaluation of the proposed framework in datasets and simulation, including performance, communication and scalability, is conducted. The source code and custom datasets are open to the public.

\section{Related Work}
\label{sec:related_work}
\subsection{LiDAR Inertial Odometry}
LiDAR odometry is typically required to operate in real-time. One of the renowned feature-based algorithms in this domain is LOAM \cite{loam:2014:zhang}, which extracts edge and plane features based on local smoothness.
Several works maintain a framework similar to LOAM, such as LIO-SAM \cite{lio_sam:2020:shan}, LRIO \cite{Nguyen_lrio}, and LiLi-OM \cite{lili-om:2021:li}.
These approaches incorporate IMU to compensate for motion distortion and provide an initial estimate for scan matching.
However, achieving real-time performance often comes at the cost of maintaining only a small local map, which can lead to drift over time. Consequently, a map refining process, such as factor graph optimization, becomes necessary.
FAST-LIO2 \cite{fast_lio2:2022:xu} introduces an ESIKF and incremental kd-tree to enhance computational efficiency, enabling direct and real-time scan-to-map registration. 
Compared to the aforementioned feature-based methods, FAST-LIO2 offers improved odometry accuracy while eliminating the need for parameter tuning for feature extraction.
Nevertheless, maintaining the incremental map can be challenging in multi-robot scenarios due to the introduction of numerous inter-robot measurements. 
LION \cite{lion:2021:tagliabue} also directly registers raw points to the map, which is very similar to the idea of GICP \cite{gicp:2009:segal} and Normal Distribution Transformation (NDT) \cite{3D-trans:2009:magnusson}. 
However, the inefficiencies of GICP \cite{gicp:2009:segal} and the limited stability of NDT are issues that demand attention.
To address these challenges, Fast-GICP \cite{koide2021vgicp} adopts the multithread technology to alleviate the computational load. Our work inherits this approach, aiming to improve efficiency and stability in front-end odometry.

\subsection{Multi-robot Localization and Mapping}
Estimating the relative pose between robots is a critical task, primarily aimed at compensating for odometry drift without the reliance on external positioning infrastructure. 
Using vision detection \cite{nguyen2020vision} has shown the capability to achieve centimeter-level accuracy in determining relative poses.
Nevertheless, it suffers from limitations related to onboard camera visibility and restricted field-of-view (FoV), which can constrain the trajectories and formation between robots.
An efficient and widely used approach involves place recognition through global or object-based features.
Visual-based collaborative Simultaneous Localization and Mapping (CSLAM) systems \cite{schmuck2021_covins, ouyang2021collaborative, tian2022kimera} share the binary descriptor (e.g., ORB) of the keyframes and employ a bag-of-words (BoWs) approach to find multi-robot loop closures.
Similarly, laser-based CSLAM systems \cite{dube2020_segmap, huang2021_discoslam, lajoie2023swarm} extract global features for loop closure detection.
LAMP \cite{Chang2022LAMP2A} use TEASER \cite{yang2020teaser} and GICP algorithm \cite{gicp:2009:segal} for relative transform estimation.
These approaches rely on having sufficient overlapping features within the robots' visual range, which can restrict swarm formation and orientation, rendering them most effective in feature-rich environments.
A different approach, presented in \cite{zhou2021online, nguyen2021viral, Nguyen_lrio}, leverages UWB measurements with a fixed anchor to eliminate odometry drift. While this strategy demonstrates promise, it necessitates additional infrastructure deployment, limiting its practicality in real-world scenarios.
Moreover, other works \cite{ziegler2021distributed,queralta2022vio,9350155,omni_swarm:2022:xu} fuse odometry, particularly VIO, with UWB distance measurements to estimate relative robot states.
Notably, in the case of Omni-swarm \cite{omni_swarm:2022:xu}, vision-based detection and place recognition are also incorporated for multi-robot state estimation.
Unlike place recognition methods, the approach in \cite{ziegler2021distributed, queralta2022vio, 9350155} does not mandate trajectory overlaps, granting it greater mission flexibility.
However, achieving global consistency and robust optimization under these methods remains challenging, mainly when dealing with line-of-sight limitations inherent to distance measurements.
This work introduces lightweight global descriptors and distance measurements for efficient loop closure detection, and robust optimization ensures global consistency, making it a promising multi-robot state estimation solution in challenging environments.
\section{Relative State Estimation Framework}
\label{sec:system_architecture}
\begin{figure*}[thpb]
	\centering
	\includegraphics[width=0.85\linewidth]{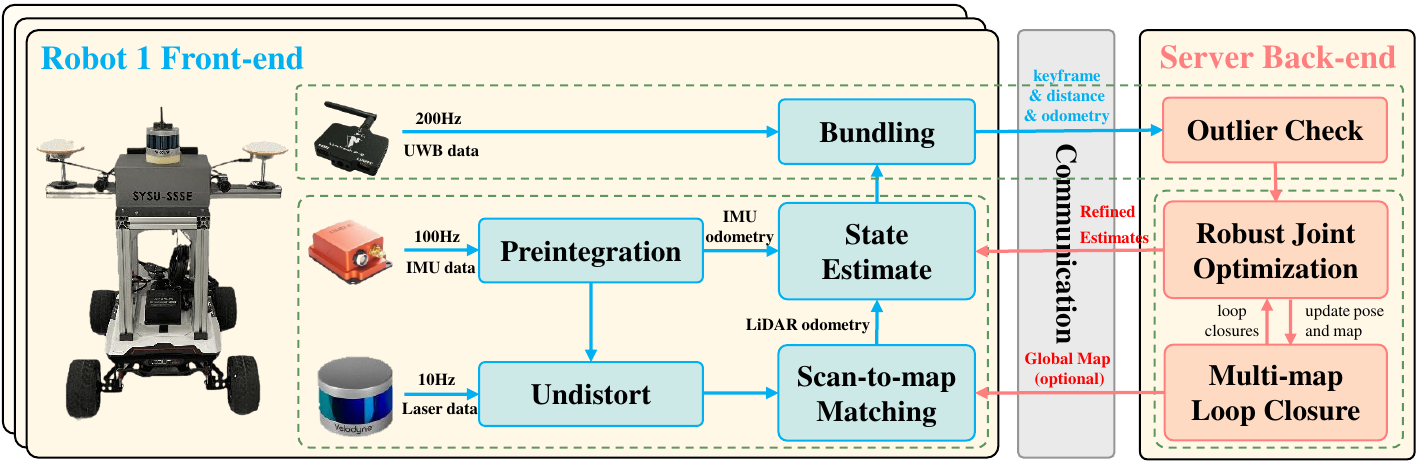}
	\caption{The system structure of CoLRIO. In the front end, the preintegration and direct scan-to-map matching odometry are further jointly optimized with fixed-lag smoothness. Keyframes generated by LIO, which include UWB measurements, are transmitted to the server. The server collects all measurements, executing robust joint optimization to achieve swarm state estimations. Finally, the estimates are returned to the front end for further processing.}
	\label{fig:architecture}
	\vspace{-0.4cm} 
\end{figure*}

The proposed system architecture comprises two primary components: a central server and a team of robots, as visually represented in Fig. \ref{fig:architecture}.
The robot with a unique identification in the team is designed to join or exit the collaborative optimization freely.
Collecting raw data from a LiDAR, an IMU, and a UWB radio, each robot performs a LIO algorithm to estimate its state.
In our implementation, we employ a tailored adaptation of Fast-GICP method for the scan-to-map matching process, finding the point's correspondence with the help of its intensity.
The odometry utilizes a fixed-lag smoother, a popular method for the visual SLAM, to solve the state estimation problem efficiently in a sliding window.

Processing every LiDAR frame for computing and incorporating factors into the back-end factor graph proves to be computationally and transmission demanding for the server.
To mitigate this, we adopt the concept of keyframe selection for real-time nonlinear optimization, striking a balance between map density and memory consumption by maintaining a relatively sparse factor graph.
Keyframes are succinctly described using the lightweight global descriptor Scan-Context++ \cite{kim2021scancontext}. 
These keyframes, relative odometry, and distance measurements are transmitted to the server. 
The Multi-map Loop Closure module detects overlaps in the trajectories of robots and reports loop closure factors.
The distance measurements are further validated against odometry poses to identify and reject outliers.
Odometry, distance, and loop closure factors perform robust joint optimization utilizing Graduated Non-Convexity (GNC) \cite{yang2020gnc} to obtain the trajectory estimates.
The process for generating the above factors is described in the following section. 

\subsection{Single-robot Front-end}
\subsubsection{IMU Preintegration Factor}
The IMU preintegration factor deals with raw IMU measurements, encompassing angular velocity $\tilde{\mathbf{w}}$ and linear acceleration $\tilde{\mathbf{a}}$ at time $t$. The equations governing these measurements are as follows:
\begin{equation}
\begin{gathered}
\tilde{\mathbf{w}}_{t} = \mathbf{w}_{t}+\mathbf{b}^{\mathbf{w}}_{t}+\mathbf{n}^{\mathbf{w}}, \ \ \ \mathbf{b}^{\mathbf{w}}_{t} = \mathbf{n}^{\mathbf{b}^{\mathbf{w}}} \\
\tilde{\mathbf{a}}_{t} = \mathbf{R}_{t}^{T}(\mathbf{a}_{t}-\mathbf{g}) +\mathbf{b}^{\mathbf{a}}_{t}+\mathbf{n}^{\mathbf{a}}, \ \ \ \mathbf{b}^{\mathbf{a}}_{t} = \mathbf{n}^{\mathbf{b}^{\mathbf{a}}}
\end{gathered}
\end{equation}
where $\mathbf{w}$ and $\mathbf{a}$ denote the true angular velocity and linear acceleration, $\mathbf{n}^{\mathbf{w}}$ and $\mathbf{n}^{\mathbf{a}}$ are zero-mean Gaussian white noise. The gyroscope bias $\mathbf{b}^{\mathbf{w}}$ and accelerometer bias $\mathbf{b}^{\mathbf{a}}$ are modeled as random walks, driving by the white Gaussian noises $\mathbf{n}^{\mathbf{b}^{\mathbf{w}}}$ and $\mathbf{n}^{\mathbf{b}^{\mathbf{a}}}$, respectively. 
We compute the relative body motion between two timestamps, similar to the IMU preintegration method introduced in \cite{forster2016imupreintegration}. With a time offset $\Delta t$ between two frames, the IMU preintegration factor $\mathbf{r}^{I}$ can be expressed as:
\begin{equation}
\mathbf{r}_{t_i,t_j}^{I} = \begin{pmatrix}
\Delta\hat{\mathbf{R}}_{t_i, t_j}^{T}(\mathbf{R}_{t_i}^{T}\mathbf{R}_{t_j}) \\
\mathbf{R}_{t_i}^{T}(\mathbf{v}_{t_j}-\mathbf{v}_{t_i}-\mathbf{g}\Delta t) - \Delta \hat{\mathbf{v}}_{t_i, t_j} \\
\mathbf{R}_{t_i}^{T}(\mathbf{p}_{t_j}-\mathbf{p}_{t_i}-\mathbf{v}_{t_i}\Delta t -\frac{1}{2}\mathbf{g}\Delta t^2) - \Delta \hat{\mathbf{p}}_{t_i, t_j}
\end{pmatrix},
\end{equation}
where $\hat{\mathbf{R}}$, $\hat{\mathbf{p}}$, and $\hat{\mathbf{v}}$ indicates the estimated rotation, translation, and velocity.

\subsubsection{LiDAR Odometry Factor}
Suppose a new LiDAR scan $\mathbf{S}$ arrives at time $t$. The previously preintegrated IMU data prior to time $t$ can be employed to mitigate distortions in the scan caused by motion. It is worth noting that this new scan $\mathbf{S}$ is represented in the body frame. The process of generating a LiDAR odometry factor is described as follows:

\textbf{Voxel map maintaining}.
Rather than optimizing the transformation between consecutive scans, we transform historical scans into the map frame using corresponding poses and compute the scan-to-map transformation for each scan. Given the high frequency of map reconstruction caused by global estimate refining, we employ a classic kd-tree-based map instead of an incremental map. Additionally, to save computational resources, keyframes are added to strike a balance between map density and memory consumption. In our work, the criteria for adding a new keyframe are set at position and rotation change thresholds of 1 m and 0.2 rad, respectively. We downsample the map at a resolution of 0.4 meters to manage computational resources effectively.


\textbf{Scan-to-map matching}.
We match a newly arrived scan $\mathbf{S}$ to the map $\mathbf{M}$ via raw point registration using the Fast-GICP method, chosen for its computational efficiency and robustness across various challenging environments. For rapid convergence, we use the predicted robot motion from IMU preintegration as the initial transformation for scan-to-map matching.
For each point $\mathbf{n}$ in scan $\mathbf{S}$, we identify its corresponding point $\mathbf{m}$ in map $\mathbf{M}$ based on both position and intensity.
Specifically, inspired by the two-stage point neighbor search approach proposed in \cite{jung2021intensity}, we first perform a position-based k-nearest-neighbor search. Subsequently, an intensity-based k-nearest-neighbor search is conducted using the neighbor points obtained from the previous step. This dual-stage process allows us to further gather points with similar geometry and materials. Furthermore, we compute the covariance matrix $\Sigma$ of the corresponding point using the neighbor points.
The transformation error between the scan $\mathbf{S}$ and the map $\mathbf{M}$ can be computed as follows:
\begin{equation}
\label{eq:transformation error}
e = \frac{1}{N}\sum_{i=1}^{N}w_{i}\lVert \mathbf{R}\cdot \mathbf{n}_i+\mathbf{p}-\mathbf{m}_i\lVert^2,
\end{equation}
where $N$ is the number of the corresponding points set, $\mathbf{R}$ and $\mathbf{p}$ are the rotation matrix and translation vector that transform the source point cloud to align with the target point cloud, $\mathbf{m}_i$ and $\mathbf{n}_i$ are the $i$-th of corresponding points, respectively, and $w_i = (\Sigma_{\mathbf{m}_i}+\mathbf{T} \Sigma_{\mathbf{n}_i} \mathbf{T}^T)^{-1}$ is the weight based on the transformation $\mathbf{T} = [\mathbf{R}|\mathbf{p}]$ and covariance matrix $\Sigma$ of the points.
We compute the final transformation $\mathbf{T}_f$, minimizing the transformation error $e$.
Subsequently, we obtain the relative transformation $\Delta\mathbf{X}_{t_i, t_j}=\mathbf{X}_{t_i}^{-1}\mathbf{X}_{t_j}=\mathbf{X}_{t_i}^{-1}\mathbf{T}_f$ between robot state $\mathbf{X}_{t_i}$ and $\mathbf{X}_{t_{j}}$, with odometry measurements modeled with Gaussian noise, denoted as $\mathbf{Z}^{O}_{t_i,t_j} = \Delta\mathbf{X}_{t_i, t_j} + \mathcal{\eta}(0,\Sigma_{O})$.
Resulting in the following odometry factor $\mathbf{r}^{O}$:
\begin{equation}
\mathbf{r}^{O}_{t_i,t_j} = (\mathbf{Z}^{O}_{t_i,t_j})^{-1} (\hat{\mathbf{X}}_{t_i})^{-1} \hat{\mathbf{X}}_{t_j},
\end{equation}
where $\hat{(\cdot)}$ indicates the estimated state.

\begin{figure}[htbp]
	\centering
	\vspace{-0.1cm} 
	\includegraphics[width=0.95\linewidth]{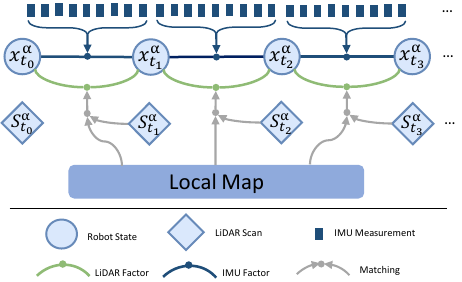}
	\caption{The structure of front-end odometry, fusing LiDAR and IMU data.}
	\label{fig:odom_posegraph}
	\vspace{-0.1cm} 
\end{figure}
As depicted in Fig. \ref{fig:odom_posegraph}, the front-end fuses different measurements in a factor graph optimization framework, including LiDAR odometry factors, IMU factors, and a prior factor (including the initial pose and the refined estimates). We formulate a Nonlinear Least-Squares (NLS) problem:

\begin{equation}
\begin{aligned}
\hat{\mathbf{X}} = \underset{\mathbf{X}}{\arg \min} \ & \bigg\{ \sum_{\mathbf{O}} \lVert \mathbf{r}^{O} \lVert^2 + \sum_{\mathbf{I}}\lVert \mathbf{r}^{I} \lVert^2 + \lVert \mathbf{r}^{P} \lVert^2 \bigg\},
\end{aligned}
\end{equation}
where $\mathbf{O}$/$\mathbf{I}$ is the set of all odometry/IMU measurements.

Since adding every odometry factor to the graph is computationally infeasible, old factors in the factor graph are marginalized to bound the problem size, which is solved through the repeated linearization of factors. Utilizing probabilistic marginalization ensures that information on the active states is well-preserved in the optimization, even as measurements and old states are removed from the sliding window. 
In the experiments, we achieve multi-source data joint optimization using an iSAM2-based fixed-lag smoother with a sliding window containing ten keyframes.


\subsection{Swarm Localization}
\subsubsection{Loop Closure Detection}
To establish additional constraints within and across the robots' trajectories, we employ loop-closure detection, utilizing the global descriptor Scan-Context++ \cite{kim2021scancontext}.
The server maintains a global database of the keyframe descriptors transmitting from robots to enable inter-robot loop detection.
New loop closures are detected by querying the database for nearest neighbor candidates, with the candidates subjected to scan-to-map matching using Fast-GICP.
Similar to equation (\ref{eq:transformation error}), the relative pose of robot $\alpha$ and $\beta$, indicated by $_{{\alpha}}\mathbf{X}_{t_i}$ and $_{\beta}\mathbf{X}_{t_m}$, in case of loop detection between different robots or within the same trajectory, respectively, is optimized by minimizing the transformation error.
The loop closure is accepted if the fitness score does not exceed the threshold. Subsequently, we obtain the relative transformation $_{\alpha,\beta}\mathbf{Z}_{t_i,t_m}$ between $_{\alpha}\mathbf{X}_{t_i}$ and $_{\beta}\mathbf{X}_{t_m}$, and the loop closure measurements modeled with a Gaussian noise, denoted as
$_{\alpha,\beta}\mathbf{Z}^{L}_{t_i,t_m} = (_{\alpha}\mathbf{X}_{t_i})^{-1} {_{\beta}}\mathbf{X}_{t_m} + \mathcal{\eta}(0,\Sigma_{L})$.
Resulting in the following loop closure factor $\mathbf{r}^{L}$:
\begin{equation}
_{\alpha,\beta}\mathbf{r}^{L}_{t_i,t_m} = (_{\alpha,\beta}\mathbf{Z}^{L}_{t_i,t_m})^{-1} ({_\alpha}\hat{\mathbf{X}}_{t_i})^{-1} {_\beta}\hat{\mathbf{X}}_{t_m}.
\end{equation}

\subsubsection{Pairwise Distance Measurement}

From the UWB modules, the relative distances between each pair of the UWB nodes can be obtained, represented as the UWB distance measurement $\mathbf{Z}^{U}_t$ at time $t$, modeled with Gaussian noise, denoted as
$_{\alpha,\beta}\mathbf{Z}^{U}_{t} = \lVert {_\alpha}\mathbf{p}_{t} - {_\beta}\mathbf{p}_{t} \lVert^2 + \mathcal{\eta}(0,\Sigma_{U})$.
The occlusion of obstacles often results in significant outliers in the UWB measurements.
To assess the consistency of the distance measurements with the IMU odometry produced by the front end, we consider any measurement that exceeds the distance outlier threshold, denoted as $_{\alpha,\beta}\mathbf{Z}^{U}_{t} - ({_\alpha}\hat{\mathbf{p}}_{t} - {_\beta}\hat{\mathbf{p}}_{t}) < \sigma$ as an outlier. Note that the position $_{\alpha}\hat{\mathbf{p}}_{t}$ and $_{\beta}\hat{\mathbf{p}}_{t}$ of IMU odometry are in the same frame. We set the distance outlier threshold to 0.2 m, which is twice the expected error of the UWB measurements. The relative distance factor $\mathbf{r}^{uwb}$ as follows:
\begin{equation}
_{\alpha,\beta}\mathbf{r}^{U}_{t} = _{\alpha,\beta}\mathbf{Z}^{U}_{t} - \lVert _{\alpha}\hat{\mathbf{p}}_{t} - {_\beta}\hat{\mathbf{p}}_{t} \lVert^2 .
\end{equation}

\subsubsection{Global Optimization}
\begin{figure}[htbp]
	\centering
	\vspace{-0.1cm} 
	\includegraphics[width=0.95\linewidth]{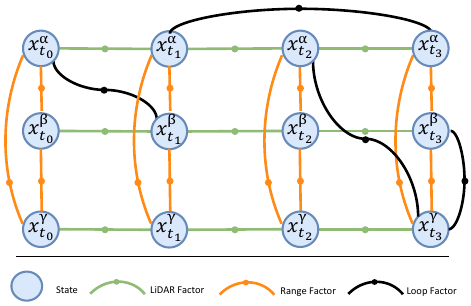}
	\caption{The structure of the CoLRIO back-end. LiDAR odometry, range, and loop closure factors are introduced to global optimization.}
	\label{fig:multirobot pose graph}
	\vspace{-0.1cm} 
\end{figure}
As illustrated in Fig. \ref{fig:multirobot pose graph}, the factors that need to be solved consists of the above LiDAR odometry factors $\mathbf{r}^{O}$, relative distance factors $\mathbf{r}^{U}$, loop closure factors $\mathbf{r}^{L}$ and a prior factor $\mathbf{r}^{P}$ (the initial pose of one of the robots). We formulate the following NLS problem:
\begin{equation}
\begin{aligned}
\hat{\mathbf{X}} = \underset{\mathbf{X}}{\arg \min} & \bigg\{ \sum_{\mathcal{O}} \lVert \mathbf{r}^{O} \lVert^2 + \sum_{\mathcal{U}}\lVert \mathbf{r}^{U} \lVert^2 + \sum_{\mathcal{L}}\lVert \mathbf{r}^{L} \lVert^2 + \lVert \mathbf{r}^{P} \lVert^2 \bigg\},
\end{aligned}
\end{equation}
where $\mathcal{O}$/$\mathcal{U}$/$\mathcal{L}$ is the set of all key odometry/distance/loop closure measurements.
To enhance the robustness and reliability of our server back-end, we have seamlessly integrated two outlier rejection mechanisms: Pairwise Consistent Measurement set maximization (PCM) \cite{mangelson2018pcm} and GNC \cite{yang2020gnc}. PCM scrutinizes inter-robot loop closures for consistency with other odometry. Meanwhile, GNC works alongside Levenberg-Marquardt to execute an outlier-robust factor graph optimization, providing trajectory estimates and making inlier/outlier determinations on the inter-robot (loops and distances) measurements not rejected by PCM. Both PCM and GNC are implemented with the GTSAM library.
\section{Experiments}
\label{sec:experiment}

\subsubsection{Setup}
All experimental results presented in this paper were derived from performance comparisons conducted on five indoor and outdoor sequences of the S3E dataset \cite{feng2022s3e} and six datasets collected on SYSU's eastern campus. We leverage Robot Operating System 2 (ROS2) \cite{ros2:2022:Macenski} and a PC equipped with an Intel i7-10700K CPU and 32GB DDR4 RAM for these experiments. Table \ref{tab:dataset infomation} summarizes the critical characteristics of the datasets utilized in our subsequent experiments.
Inheriting the hardware of three ground vehicles in the S3E dataset, we added a LinkTrack P-B for inter-robot distance measurement, upgraded the original 1Hz Real-time Kinematic (RTK) system to a high-frequency 100Hz CHCNAV CGI-610 GNSS unit as the ground truth to ensure the accuracy and reliability of the ground truth data (especially elevation). Note that AMOVLAB provides some devices for ground vehicles.
Moreover, the CGI-610 served as an external trigger, enabling precise synchronization of the Xsens MTi-30 IMU and Velodyne VLP-16 Puck LiDAR systems. Additionally, it was instrumental in maintaining synchronized system time among the robotic platforms via GPS timing and synchronization.
To evaluate the trajectories, we employed ground truth to compute the Absolute Translation Error (ATE) using evo \cite{grupp2017evo} toolbox.
For indoor S3E sequences, only the initial and final positions are compared.
\begin{table}[htbp]
	\centering
	\vspace{-0.4cm} 
	\setlength{\abovecaptionskip}{-0.3cm}
	\caption{Details of S3E and Our Campus Datasets}
	\label{tab:dataset infomation}
	\begin{center}
		\tabcolsep=0.14cm
		\begin{tabular}{ccccccc}
			\toprule
			\multicolumn{2}{c}{\multirow{2}{*}{$\mathbf{Datasets}$}} & \multirow{2}{*}{$\mathbf{Sensor}$} & \multicolumn{3}{c}{$\mathbf{Trajectory\ Length}$ {[}m{]}} & \multirow{2}{*}{$\mathbf{Size }$ [GB]}\\
			&  & \multicolumn{1}{c}{} & Robot 1 & Robot 2 & Robot 3 \\
			\midrule
			\multirow{3}{*}{\rotatebox{90}{S3E}} & \textit{Square\_1} & LVI & 455.6 & 454.4 & 458.2 & 19.1 \\
			& \textit{College} & LVI & 920.5 & 995.9 & 1072.3 & 31.6\\
			& \textit{Playground\_2} & LVI & 277.7 & 319.9 & 474.0 & 6.7\\
			& \textit{Laboratory\_1} & LVI & 146.7 & 161.6 & 141.2 & 9.4\\
			& \textit{Laboratory\_2} & LVI & 213.5 & 197.9 & 160.2 & 6.7\\
			\midrule
			\multirow{6}{*}{\rotatebox{90}{Our dataset}} & \textit{Archway} & LRI & 487.4 & 569.6 & 563.8 & 8.5\\
			& \textit{Rotation} & LRI & 84.7 & 91.3 & 110.7 & 1.8\\
			& \textit{Small\_Loop} & LRI & 521.6 & 523.9 & 520.3 & 9.3\\
			& \textit{Large\_Loop} & LRI & 1938.6 & 1934.2 & 1950.1 & 30.9 \\
			& \textit{Tunnel} & LRI & 521.9 & 502.4 & 501.1 & 8.2 \\
			& \textit{College} & LRI & 983.3 & 967.6 & 986.9 & 17.5 \\
			\bottomrule
		\end{tabular}
	\end{center}
	\footnotesize{$\dagger$ LVI indicates LiDAR-Visual-Inertial and LRI indicates LiDAR-Ranging-Inertial.}
	\vspace{-0.2cm} 
\end{table}
\begin{table*}[t]
	\centering
	\setlength{\abovecaptionskip}{-0.3cm}
	\caption{Absolute Translation Error (meter) of the S3E Sequences and Our Campus Datasets}
	\label{tab:gnss dataset evaluation}
	\begin{center}
		\tabcolsep=0.08cm
		\begin{tabular}{c|c|c|c|c|c|c|c|c|c|c|c}
			\toprule
			\multicolumn{1}{c|}{\multirow{2}{*}{\diagbox{Configure}{Datasets}}} & \multicolumn{6}{c|}{Our campus dataset} & \multicolumn{5}{c}{S3E dataset} \\
			
			\multicolumn{1}{c|}{} & \textit{Archway} & \textit{Rotation} & \textit{Small\_Loop} & \textit{Large\_Loop} & \textit{Tunnel} & \textit{College} & \textit{Playground\_2} & \textit{Square\_1} & \textit{College} & \textit{Laboratory\_1} & \textit{Laboratory\_2}\\
			
			\midrule
			LIO-SAM & 54.22 & 5.71 & 1.01 & 131.52 & 3.85 & 49.24 & 24.81 & 0.94 & 3.05 & 3.99 & 11.06 \\
			
			FAST-LIO2 & 4.55 & 0.35 & 1.83 & 35.00 & 4.35  & 5.58 & 20.72 & 3.00 & 8.30 & 8.30 & 9.13  \\
			
			our front-end & 1.63 & $\mathbf{0.26}$ & 0.92 & 6.16 & $\mathbf{3.53}$ & 0.81 & $\mathbf{0.48}$ & $\mathbf{0.73}$ & 1.01 & $\mathbf{0.46}$ & 0.57 \\
			
			\midrule
			
			Swarm-SLAM & 6.15 & 3.72 & 3.20 & 22.42 & 10.54  & 7.93 & 1.45 & 4.20  & 3.57 & 4.76 & 8.97 \\
			
			DiSCo-SLAM & 7.43 & 5.79 & 1.03 & 23.38 & 7.71 & 12.02 & 3.82 & 1.03 & 1.76 & 1.20 & 2.33 \\
			
			
			
			
			CoLRIO w/o UWB & 1.64 & $\mathbf{0.26}$ & 0.94 & 1.97 & 5.07  & 1.08 & 0.63 & 0.75 & $\mathbf{0.98}$ & $\mathbf{0.46}$ & $\mathbf{0.18}$ \\
			
			our CoLRIO & $\mathbf{1.59}$ & $\mathbf{0.26}$ & $\mathbf{0.88}$ & $\mathbf{1.47}$ & 3.54  & $\mathbf{0.77}$ & 0.63 & 0.75 & $\mathbf{0.98}$ & $\mathbf{0.46}$ & $\mathbf{0.18}$ \\

			\bottomrule
		\end{tabular}
	\end{center}
	\vspace{-0.6cm} 
\end{table*}

\subsubsection{Dataset Experiments}

\begin{figure}[htbp]
	
	\setlength{\abovecaptionskip}{-0.05cm}
	\centering
	\subfigure[\textit{Large\_Loop}]
	{
		\label{fig:trajetory:a}
		{\includegraphics[width=0.465\linewidth]{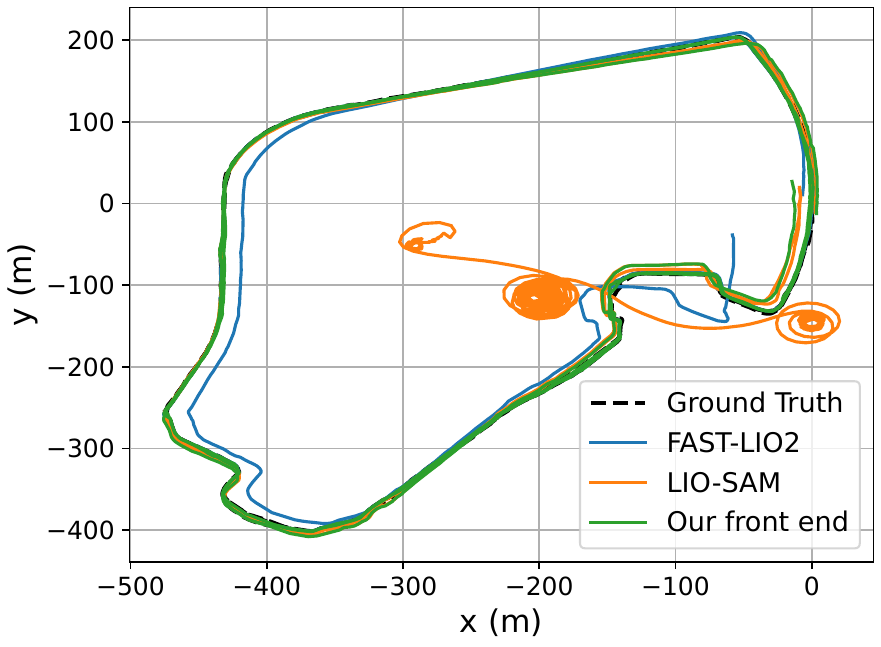}}
	}
	\subfigure[\textit{S3E\_College}]
	{
		\label{fig:trajetory:b}
		{\includegraphics[width=0.465\linewidth]{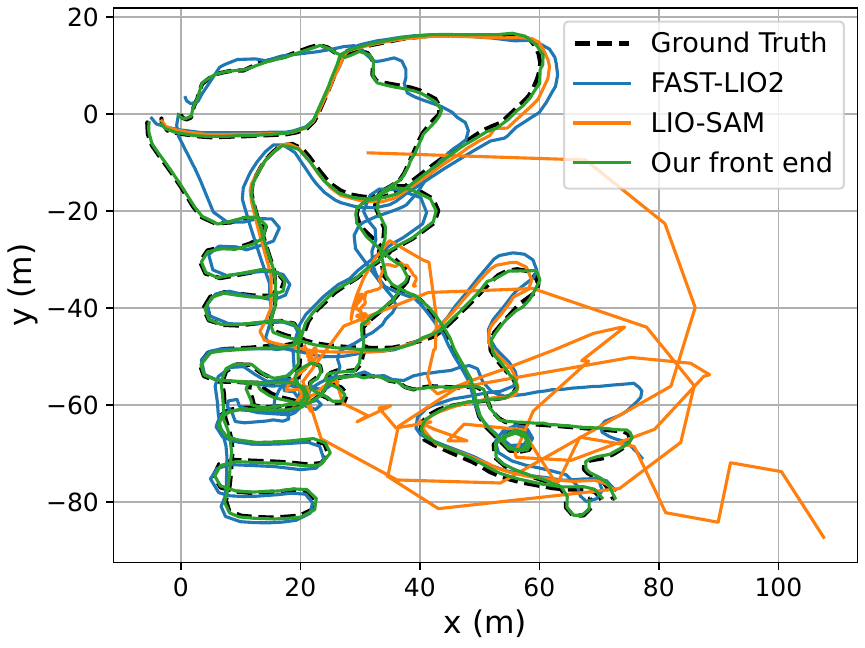}}
	}
	\subfigure[\textit{Rotation}]
	{
		\label{fig:trajetory:c}
		{\includegraphics[width=0.465\linewidth]{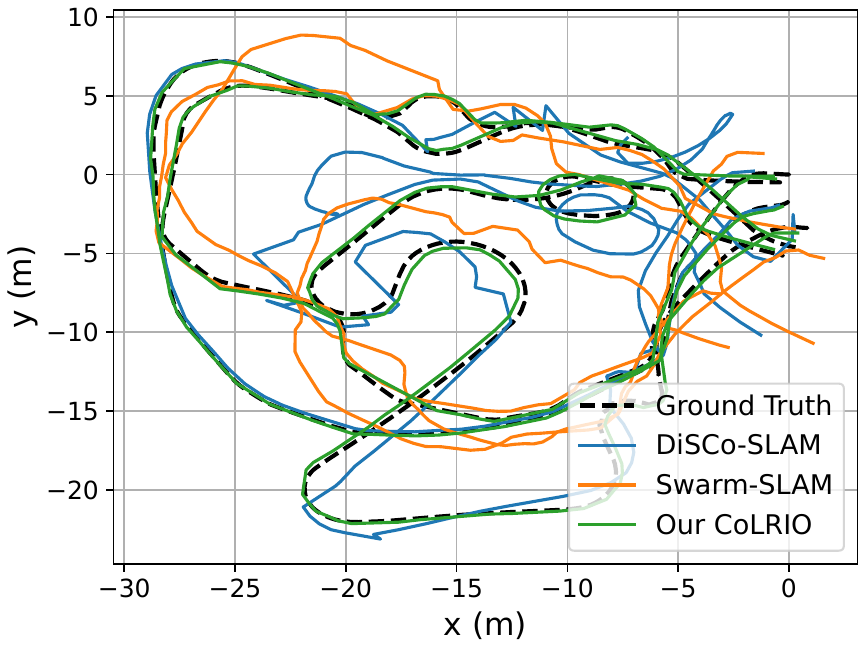}}
	}
	\subfigure[\textit{Archway}]
	{
		\label{fig:trajetory:d}
		{\includegraphics[width=0.465\linewidth]{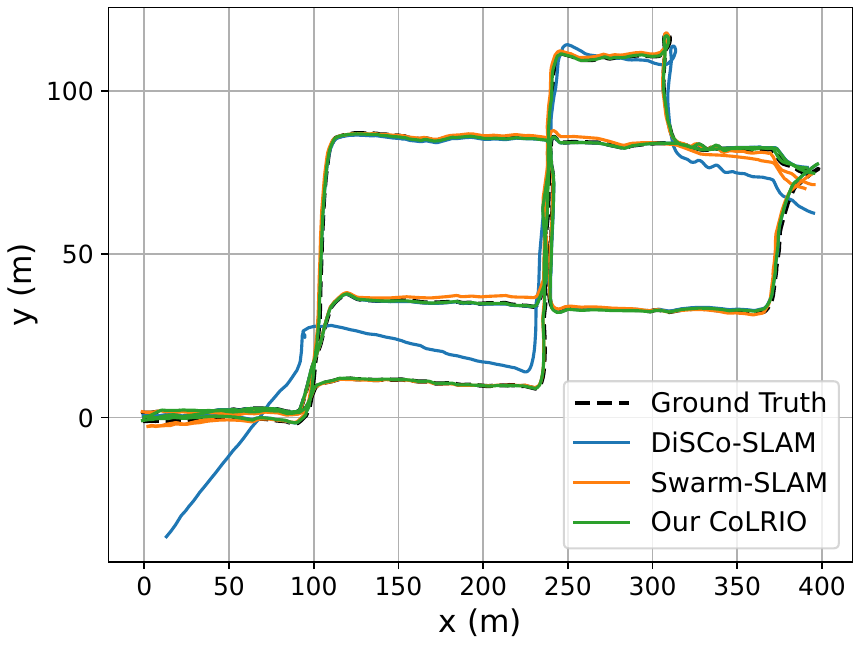}}
	}
	\caption{Different methods' trajectory estimates on various dataset sequences compared with ground truth. }
	\label{fig:trajetory}
	\vspace{-0.3cm} 
\end{figure}

To comprehensively evaluate the accuracy and robustness of our front-end odometry, we conducted a comparative analysis, pitting the proposed CoLRIO front end against state-of-the-art odometry such as LIO-SAM and FAST-LIO2 across diverse datasets.
Note that all front-ends utilize the centralized server to detect only intra-robot loop closure to reduce odometry drift and optimize the trajectory with GNC.
The evaluation results, specifically the ATE, are presented in Table \ref{tab:gnss dataset evaluation}. Fig. \ref{fig:trajetory:a} and \ref{fig:trajetory:b} illustrate the trajectories achieved by different front ends. Notably, the \textit{Archway}, \textit{Tunnel}, and \textit{S3E\_Square} sequences do not include intra-robot loop closures.
From Table \ref{tab:gnss dataset evaluation}, it's evident that the CoLRIO front end consistently outperforms single-robot odometry methods, including LIO-SAM and FAST-LIO2, thereby demonstrating its effectiveness and stability.
Thanks to its efficient direct registration utilizing raw point clouds, the front-end odometry can seamlessly adapt to challenging environmental conditions.
This enables enhanced precision in state estimation even during rapid motion (\textit{S3E\_Playground\_2}), extensive angular rotation (\textit{Rotation}), long-distance traversal (\textit{Large\_Loop}), and indoor (\textit{S3E\_Laboratory\_1} and \textit{S3E\_Laboratory\_2}) settings.
Additionally, regarding computational cost, with 4 parallel threads for the GICP algorithm, each robot performs our front-end with a maximum CPU usage of 17.4\%, taking an average of 50.03ms for each frame in the \textit{Large\_Loop} sequence, highlighting its effectiveness with real-time performance.
In summary, our front end exhibits efficient and precise localization performance, laying the foundation for joint optimization in the robotic swarms scenarios.

In order to further assess the precision and resilience of our entire SLAM system, we performed a comparative analysis by benchmarking the proposed CoLRIO method against Swarm-SLAM and DiSCo-SLAM across various datasets.
Fig. \ref{fig:trajetory:c} and \ref{fig:trajetory:d} depict the trajectories achieved by different C-SLAM systems.
Throughout our experiments, we observed that DiSCo-SLAM and Swarm-SLAM exhibited satisfactory performance in specific sequences.
However, they encountered difficulties in merging all sub-maps across most sequences.
Based on experimental observations, the limitations of DiScO-SLAM can be attributed to deficiencies in their loop closure searching module and the lack of robustness in the front-end LIO-SAM. In contrast, Swarm-SLAM exhibits elevated CPU usage in front-end RTAB-Map and loop closure processes. Insufficient computational resources may result in non-real-time odometry localization (less than 1Hz) and significantly impact the performance of the loop closure module.
Unlike DiSCo-SLAM and Swarm-SLAM, our CoLRIO approach demonstrates very competitive performance on most sequences, showcasing its robustness for different environments.
Comparing CoLRIO to the front-end, we observed performance improvements in the \textit{Large\_Loop}, \textit{College} (characterized by long duration), and \textit{Tunnel} (marked by severe perceptual aliasing). These advancements were primarily attributed to leveraging distance measurements from UWB radio, facilitating robust optimization, and effectively eliminating outlier loop closure measurements.
In the other three sequences, all robots followed forward trajectories, eventually meeting at specific points (\textit{Archway}) or returning to their initial positions after a short journey (\textit{Small\_Loop} and \textit{Rotation}). 
Although our proposed system merged these trajectories, it could not mitigate front-end drift further due to insufficient inter-robot measurements.
In short, the proposed CoLRIO framework demonstrates its effectiveness in merging sub-maps among robotic teams and robustness in challenging environments with the help of distance data.

\subsubsection{Communication}
\begin{table}[tbp]
	\centering
	\setlength{\abovecaptionskip}{-0.2cm}
	\caption{Network Traffic of Each Robot Averaged over S3E and Campus Sequences}
	\label{tab:bandwidth}
	\begin{center}
		\begin{tabular}{c|c|c}
			\toprule
			$\mathbf{Datasets}$ & $\mathbf{Robot \rightarrow Server}$ $\pm$ Std. & $\mathbf{Server \rightarrow Robot}$ $\pm$ Std.\\
			\midrule
			\textit{Archway} & 610.31 $\pm$ 33.35 kBps & 0.67 $\pm$ 0.90 kBps\\
			\textit{Small\_Lopp} & 663.54 $\pm$ 22.56 kBps & 9.89 $\pm$ 7.23 kBps\\
			\textit{S3E\_College} & 608.40 $\pm$ 34.29 kBps & 11.91 $\pm$ 10.99 kBps\\
			\cmidrule{1-3}
			Avg. (3 Seq.) & 626.69 kBps & 8.63 kBps \\
			\midrule
		\end{tabular}
	\end{center}
	\vspace{-0.8cm} 
\end{table}
To quantify the bandwidth requirements of the proposed system, we conducted a comprehensive network traffic analysis on both the S3E and our campus datasets (using the Velodyne VLP-16). The average bandwidth utilization is presented in Table \ref{tab:bandwidth}
In this configuration, each robot communicates with the central server at a frequency ranging from 1 to 2 Hz, providing new keyframe data that includes combined odometry and distance measurements. On average, these keyframe messages possess a size of approximately 242.05 kB. Reciprocally, the server communicates with individual robots at a similar frequency range of 1-2 Hz.
As delineated in Table \ref{tab:bandwidth}, the network traffic generated by the three robots and conveyed to the server amounts to approximately 600 kBps. The LinkTrack P-B UWB module and a standard WiFi module efficiently manage this data transfer.
In contrast, the server's data traffic to each robot is notably lower in our implementation. This is primarily because only necessary global estimates are shared to facilitate the drift correction.
It is important to note that transmitted data will increase rapidly for more complex sequences, where the LIO generates a greater number of keyframes. Consequently, the system's bandwidth imposes limitations on the scalability of the robotic team, making it challenging to deploy the system in larger-scale swarms comprising hundreds of robots.

\subsubsection{Simulation}
\begin{figure}[tb]
	\centering
	\includegraphics[width=0.99\linewidth]{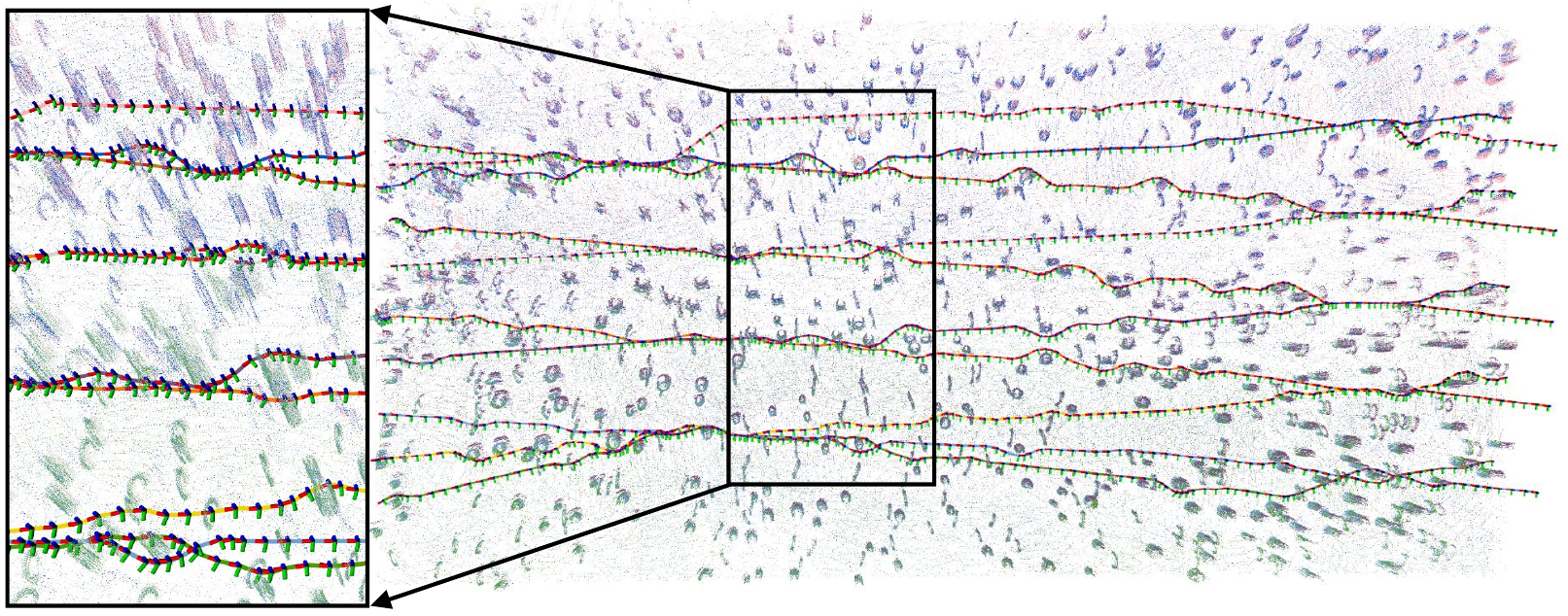}
	\caption{The reconstructed map of CoLRIO from 10 drones ﬂying across an area full of circle and cylinder obstacles.}
	\label{fig:sim10_map}
	\vspace{-0.3cm} 
\end{figure}

In this experiment, we assess the applicability of our proposed CoLRIO framework in a scenario involving a substantial team of UAVs.
For simulation purposes, we employ MARSIM \cite{kong2023marsim}, allowing us to orchestrate the flight of 10 drones within an area filled with ring and cylinder obstacles.
The trajectories followed by these drones, as depicted in Fig. \ref{fig:sim10traj}, collectively span over 100 meters and encompass the entire area.
The collaborative mapping results produced by CoLRIO are showcased in Fig. \ref{fig:sim10_map}.
The average ATE achieved by CoLRIO, when incorporating distance measurements, stands at a commendable 0.23 m. In contrast, CoLRIO's ATE, without distance measurements, registers at 0.32 m.
The experimental results show that the proposed CoLRIO framework can be extended to a certain scale robotic swarm.

\section{CONCLUSIONS}
\label{sec:conclusions}
This paper introduces CoLRIO, a robust and precise collaborative SLAM framework. CoLRIO empowers multiple robots to collaboratively generate global SLAM estimates from their concurrently collected data in real-time. Our extensive experiments underscore the accuracy and robustness of collaborative SLAM estimates, particularly in large-scale multi-agent missions, where we demonstrate its effectiveness with up to 10 robots simultaneously contributing to the system.
In our future work, we aim to extend the scalability of this system, enabling it to gracefully accommodate a multitude of agents, potentially in the hundreds, through innovative strategies such as leveraging MARSIM simulation and employing a distributed or decentralized structure. Additionally, we are committed to enhancing the system's resilience to significant outlier distance and loop measurements using robust incremental optimization.
\newpage
\addtolength{\textheight}{-12cm} 
\bibliographystyle{IEEEtran}
\bibliography{IEEEabrv,relatepaper}

\end{document}